\documentclass{article}

\usepackage{microtype}
\usepackage{graphicx}
\usepackage{subcaption}
\usepackage{booktabs} 

\usepackage{hyperref}
\usepackage{comment}
\usepackage{booktabs}
\usepackage{colortbl}
\usepackage{multirow}
\usepackage{makecell}
\usepackage{xcolor}

\definecolor{lightgray}{gray}{0.93}
\definecolor{darkgreen}{RGB}{0,130,0}

\usepackage[accepted]{icml2026}

\usepackage{amsmath}
\usepackage{amssymb}
\usepackage{mathtools}
\usepackage{amsthm}
\usepackage{comment}

\usepackage[capitalize,noabbrev]{cleveref}

\theoremstyle{plain}

\theoremstyle{definition}

\theoremstyle{remark}

\usepackage[textsize=tiny]{todonotes}

\icmltitlerunning{SwiftVLM: Efficient Vision-Language Model Inference via Cross-Layer Token Bypass}

\begin{document}

\twocolumn[
  \icmltitle{SwiftVLM: Efficient Vision-Language Model Inference via\\Cross-Layer Token Bypass}



  \icmlsetsymbol{equal}{*}

  \begin{icmlauthorlist}
    \icmlauthor{Chen Qian}{equal,yyy}
    \icmlauthor{Xinran Yu}{equal,yyy}
    \icmlauthor{Danyang Li}{yyy}
    \icmlauthor{Guoxuan Chi}{yyy}
    \icmlauthor{Zheng Yang}{yyy}
    \icmlauthor{Qiang Ma}{yyy}
    \icmlauthor{Xin Miao}{yyy}

  \end{icmlauthorlist}

  \icmlaffiliation{yyy}{Tsinghua University, Beijing, China}

  \icmlcorrespondingauthor{Chen Qian}{chen.cronus.qian@gmail.com}


  \vskip 0.3in
]



\printAffiliationsAndNotice{}  

\setlength{\textfloatsep}{5pt}
\setlength{\intextsep}{6pt}
\setlength{\floatsep}{4pt}

\begin{abstract}
Visual token pruning is a promising approach for reducing the computational cost of vision–language models (VLMs), and existing methods often rely on early pruning decisions to improve efficiency.
While effective on coarse-grained reasoning tasks, they suffer from significant performance degradation on tasks requiring fine-grained visual details.
Through layer-wise analysis, we reveal substantial discrepancies in visual token importance across layers, showing that tokens deemed unimportant at shallow layers can later become highly relevant for text-conditioned reasoning.
To avoid irreversible critical information loss caused by premature pruning, we introduce a new pruning paradigm, termed bypass, which preserves unselected visual tokens and forwards them to subsequent pruning stages for re-evaluation.
Building on this paradigm, we propose SwiftVLM, a simple and training-free method that performs pruning at model-specific layers with strong visual token selection capability, while enabling independent pruning decisions across layers.
Experiments across multiple VLMs and benchmarks demonstrate that SwiftVLM consistently outperforms existing pruning strategies, achieving superior accuracy–efficiency trade-offs and more faithful  visual token selection behavior.
\end{abstract}

\section{Introduction}

Vision–Language Models (VLMs)~\cite{team2024gemini, chen2024internvl, alayrac2022flamingo} have rapidly advanced in recent years and emerged as a central paradigm in multimodal learning. These models integrate a visual encoder with a large language model (LLM)~\cite{grattafiori2024llama, achiam2023gpt} through a cross-modal fusion module, enabling strong performance across a wide range of vision–language tasks~\cite{gao2025vla, lin2025healthgpt, yang2025re, wang2025sharpness}. In practice, visual inputs are processed by generating a large number of visual tokens. However, only a small subset of these tokens is critical for text-conditioned reasoning, with the remainder largely increasing latency and computational overhead.

To reduce the number of visual tokens, prior studies adopt token merging strategies, such as ToMe~\cite{bolya2022token}, Qwen-VL~\cite{bai2025qwen2}, and VisionZip~\cite{yang2025visionzip}.
These methods aggregate visual features based on feature similarity or spatial proximity. While these approaches improve inference efficiency, such compression degrades fine-grained visual details, especially for precise localization tasks.

\begin{figure}[t]

  \begin{center}
    \centerline{\includegraphics[width=\columnwidth]{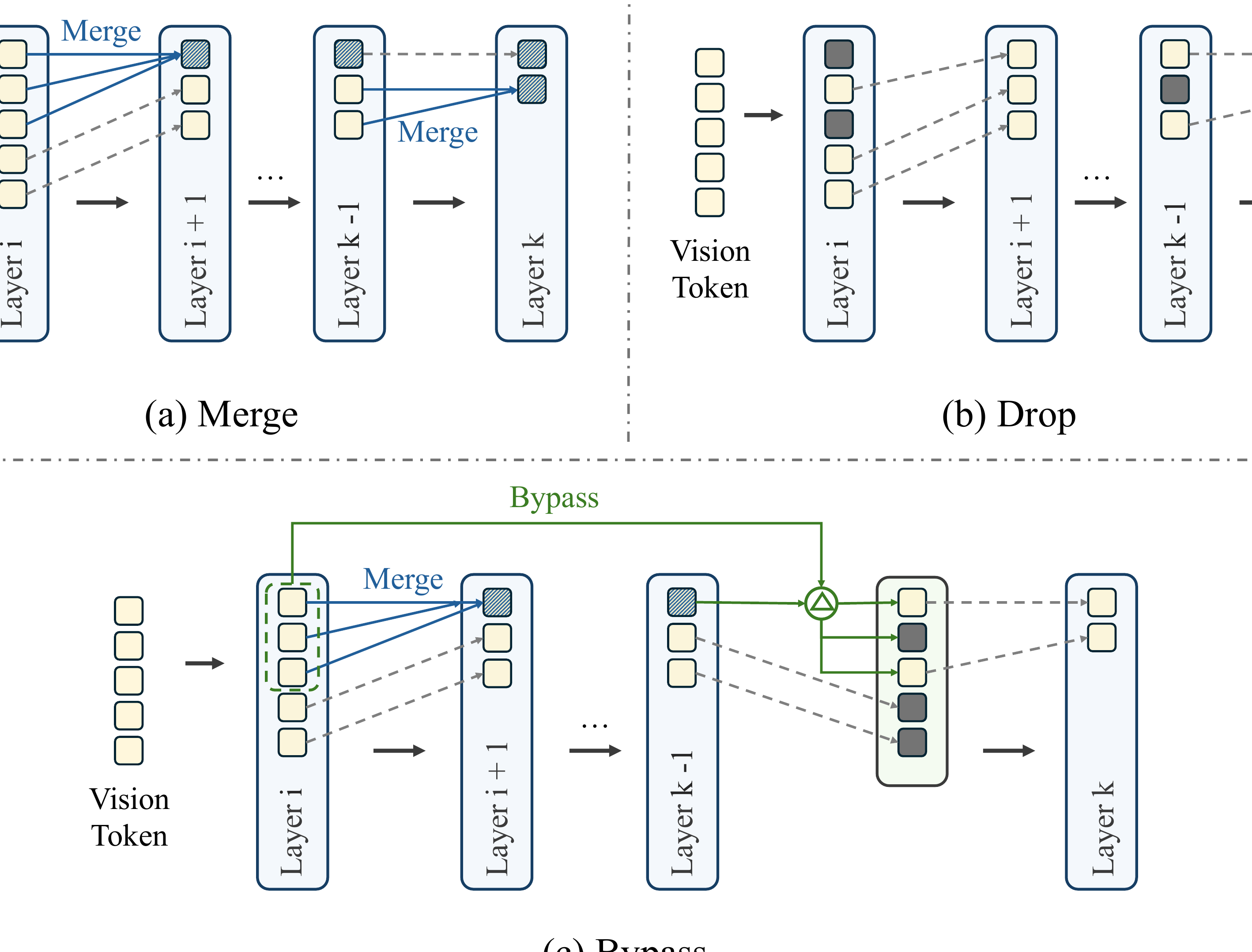}}
    \caption{
      \textbf{Comparison of visual token pruning strategies in VLMs.} (a)–(b) Existing approaches suffer from irreversible loss of critical visual information once tokens are merged or dropped in shallow layers.
      (c) We propose Bypass, a pruning strategy that restores previously merged tokens via token alignment. Bypass provides critical visual tokens with an opportunity to be reconsidered at deeper layers with stronger token selection capability.}
    \label{fig:intro-diff}
  \end{center}
\end{figure}

\begin{figure}[t]
  \begin{center}
    \centerline{\includegraphics[width=0.8\columnwidth]{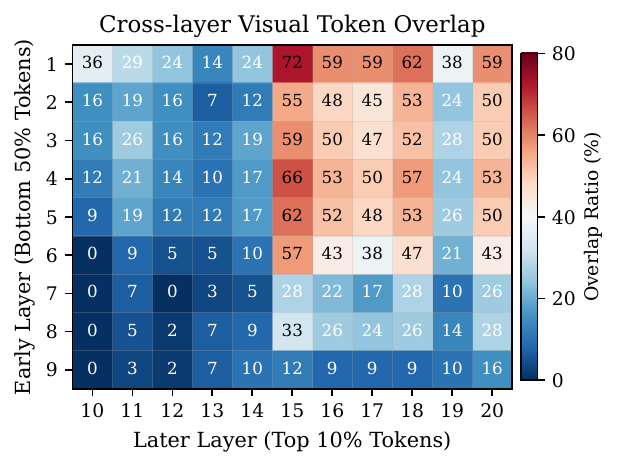}}
    \caption{
      \textbf{Layer-wise variation in visual token ranking.} For a representative TextVQA example, we report the overlap ratio between the bottom-ranked 50\% of visual tokens selected at layers 1–9 and the top-ranked 10\% selected at layers 10–20 of LLaVA.}
    \label{fig:intro-heatmap}
  \end{center}
\end{figure}


Another line of work leverages text-to-vision (T–V) attention in VLMs to rank visual tokens and dynamically drop low-ranked ones, as illustrated in Fig.\ref{fig:intro-diff}(b).
FastV~\cite{chen2024image} observes that T–V attention becomes highly concentrated on a small subset of visual tokens from the third layer onward, and thus aggressively drops low-ranked ones in a shallow layer.
PDrop~\cite{xing2024pyramiddrop} further shows that aggressive pruning in early layers leads to significant performance degradation, whereas the impact becomes less severe in deeper layers, motivating a progressive dropping strategy.
This principle is subsequently adopted by works such as SparseVLM~\cite{zhang2024sparsevlm} and FEATHER~\cite{endo2025feather}. However, we find that \textbf{the importance ranking of visual tokens varies across layers}.



\begin{figure}[t]
  \begin{center}
    \centerline{\includegraphics[width=\columnwidth]{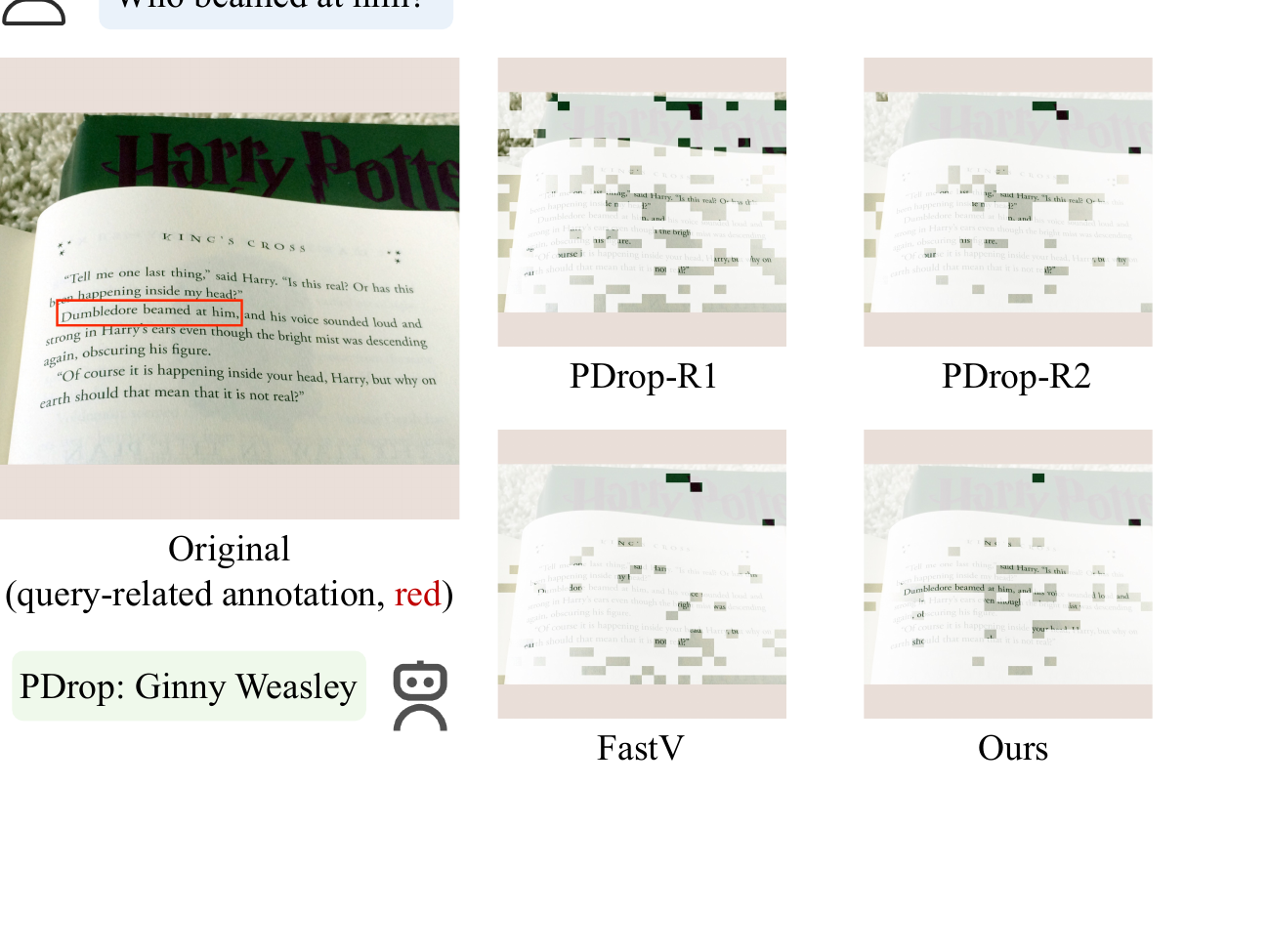}}
    \caption{
      \textbf{Comparison of results from different pruning methods.} FastV applies aggressive early-layer pruning, whereas PDrop adopts progressive pruning. Both drop the visual token containing “NASRI”, leading to incorrect answers. SwiftVLM preserves the query-relevant token at the final stage and answers correctly.}
    \label{fig:intro-visual}
  \end{center}
\end{figure}

 As illustrated in Fig.\ref{fig:intro-heatmap}, we report the overlap ratio on a TextVQA~\cite{singh2019towards} sample between the bottom 50\% visual tokens selected by early layers (layers 1–9) and the top 10\% visual tokens selected by later layers (layers 10–20) of LLaVA-1.5-7B~\cite{liu2024improved}. We observe that visual tokens deemed unimportant and dropped in early layers can become highly important in deeper layers. 

While existing methods perform early-layer pruning to improve efficiency, prematurely dropping task-relevant visual tokens can hinder subsequent reasoning. As shown in Fig.\ref{fig:intro-visual}, methods such as FastV and PDrop force deeper layers to reason over incomplete visual evidence, often resulting in incorrect answers.

Based on these observations, we propose a third pruning paradigm, termed \textbf{bypass}.
As illustrated in Fig.\ref{fig:intro-diff}(c), at the first pruning layer, bottom-ranked visual tokens are not immediately discarded. Instead, they are fully preserved and forwarded directly to the next pruning layer for re-ranking of their importance.
Meanwhile, these bottom visual tokens are merged according to feature similarity. The merged visual tokens then participate in subsequent inference.

At the following pruning layer, we derive a hidden-state offset from the merged visual tokens and use it to adjust the bypassed bottom-ranked tokens, aligning them with text tokens in the current representation space. These corrected tokens are then reintroduced for joint re-evaluation.

This design preserves the complete visual information while allowing each pruning layer to independently assess token importance, thereby avoiding irreversible critical information loss caused by premature pruning in early layers.

\begin{figure}[t]
  \begin{center}
    \centerline{\includegraphics[width=\columnwidth]{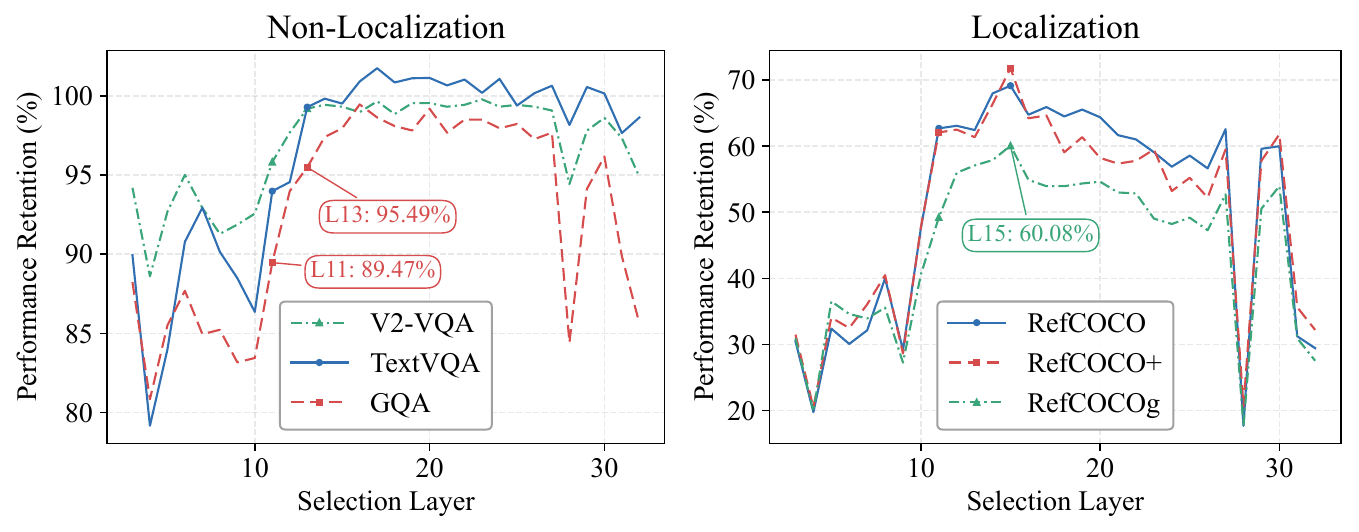}}
    \caption{
      \textbf{Non-monotonic layer-wise capability for visual token selection.} Across tasks and datasets, we record the layer-wise top 20\% visual tokens of the vanilla model and re-evaluate it by retaining all tokens in layers 1–2 and only the layer-specific top 20\% from layer 3 onward. Performance is reported relative to the vanilla baseline.}
    \label{fig:intro-select}
  \end{center}
\end{figure}

Furthermore, to determine the pruning layers used for token selection, we conduct a comprehensive layer-wise analysis across two task categories and six benchmark datasets. We first run the vanilla model and record, at each layer, the indices of the top 20\% visual tokens selected based on T–V attention. Using the same set of evaluation samples, we then re-run the model while retaining all visual tokens in the first two layers and keeping only the layer-specific top 20\% visual tokens from the third layer onward. The layer-wise results are reported in Fig.\ref{fig:intro-select}.

The results indicate that the ability to identify important visual tokens varies across layers and is \textbf{not monotonically increasing with depth}. Moreover, intermediate layers generally exhibiting stronger selection capability. Accordingly, we formulate the pruning-layer selection problem as a dynamic programming task, enforcing a monotonic increase in selection capability across the chosen pruning layers.

Based on these two observations, we propose \textbf{SwiftVLM}, a training-free method that performs pruning at layers with strong selection capability while ensuring independent pruning decisions at each stage.


We first identify model-specific optimal pruning layers (e.g., $i$ and $k$ in Fig.\ref{fig:intro-diff}(c)) and fix them for evaluation at test time. After visual token pruning at layer $i$, the unselected visual tokens are preserved and re-evaluated at layer $k$ with high selection capability.

The key contributions are summarized as follows:
\begin{itemize}
\setlength{\itemsep}{0pt}
    \item We reveal pronounced layer-wise disparities in visual token importance and propose \textbf{bypass}, a novel pruning strategy that forwards unselected visual tokens to subsequent pruning layers, enabling independent selection decisions.
    \item We reveal that the discriminative capability of layers for identifying critical visual tokens varies significantly across depth, exhibiting non-monotonic behavior.
    \item We present \textbf{SwiftVLM}, a simple yet effective training-free method that identifies high-discriminability pruning layers via dynamic programming and employs bypass to preserve fine-grained visual details while accelerating inference.
    \item Extensive experiments across two VLMs on nine benchmarks show SwiftVLM substantially outperforms existing training-free methods.
\end{itemize}
\section{Related Work}
To reduce the number of visual tokens and improve inference efficiency, existing studies~\cite{zhong2025aim, wang2025corematching, li2024llama} can be broadly classified into two categories.

\textbf{Text-agnostic.} Qwen2.5-VL~\cite{bai2025qwen2} merges each group of four neighboring visual tokens into a single token. ToMe~\cite{bolya2022token} performs similarity-based token merging between the attention and MLP blocks. VisionZip~\cite{yang2025visionzip} retains tokens with high [CLS]-attention scores and merges the remaining ones based on feature similarity, following a strategy similar to VisPruner~\cite{zhang2025beyond} and Prumerge~\cite{shang2025llava}. VoCo-LLAMA~\cite{ye2025voco} compresses visual information into a single learnable VoCo token, which is then used for subsequent cross-modal interaction.

Despite their efficiency, these methods rely solely on visual cues for token reduction, which limits their ability to preserve query-relevant visual details, particularly when the queried regions are not visually salient.

\textbf{Text-aware.} Q-Former~\cite{li2023blip} reduces visual token redundancy by training cross-modal modules that compress hundreds of visual tokens into a small set of learnable tokens. ATP-LLaVA~\cite{ye2025atp} instead introduces trainable modules within the VLM and prunes visual tokens based on importance scores derived from text–vision and vision–vision attention. Although these approaches leverage the text query to guide visual token compression or selection, they require additional trainable components, incurring extra optimization overhead.

Several training-free methods exploit the native cross-modal attention of VLMs. FastV~\cite{chen2024image} uses T-V attention to assess visual token importance and performs aggressive pruning at a shallow layer. PDrop~\cite{xing2024pyramiddrop} progressively reduces visual tokens across layers, based on the observation that pruning becomes less harmful at deeper layers. FEATHER~\cite{endo2025feather} further refines this strategy by mitigating the influence of Rotary Position Embedding (RoPE)~\cite{su2024roformer} on T-V attention, while SparseVLM~\cite{zhang2024sparsevlm} performs adaptive layer-wise pruning by estimating redundancy from the rank of the T-V attention matrix. Despite being training-free, these methods assume that tokens pruned early remain unimportant in deeper layers, which often fails in fine-grained visual reasoning, leading to performance degradation.
\section{Method}
\subsection{Preliminary: Attention in VLMs}

Let $L$ denote the total number of tokens participating in computation.
Let $h \in \mathbb{R}^{L \times d}$ denote the hidden states of all tokens. The query and key matrices are obtained via linear projections,
\begin{equation}
\mathbf{Q} = \mathbf{h} \mathbf{W}_Q, \quad
\mathbf{K} = \mathbf{h} \mathbf{W}_K.
\end{equation}
A single-head attention matrix $A \in \mathbb{R}^{L \times L}$ in a VLM is then defined as
\begin{equation}
A = \mathrm{Softmax}\!\left(\frac{\mathbf{Q}\mathbf{K}^{\top}}{\sqrt{d}}\right).
\end{equation}

VLMs adopt causal attention, under which each token is restricted to attending only to preceding tokens. As a result, the last text token attends to all input tokens. In practice, we extract its attention scores as the cross-modal component to evaluate the importance of visual tokens. Note that positional information is preserved during the pruning process.

\subsection{Pruning Layer Selection}

In this section, we focus on how to accurately select pruning layers with high discriminative capability. Note that we exclude the first two layers from our analysis, as these layers exhibit distinct characteristics compared to other layers~\cite{lad2024remarkable, kang2025your}.

For a model with $L$ layers, we first record the top $V\%$ visual tokens selected by T–V attention at each layer using the vanilla model. Keeping the text and image inputs unchanged, we then re-evaluate the model by retaining all tokens in the first two layers and only the layer-specific top $V\%$ visual tokens from the third layer onward, producing a layer-wise performance profile. This performance sequence reflects the ability of each layer to identify task-relevant visual tokens. We formulate this as:
\begin{equation}
    \left \{ x_i \right \}_{i=1}^L,\quad x_i\in \mathbb{R}.
\end{equation}
Intuitively, the progressively selected pruning layers should exhibit monotonically increasing performance in this sequence. Let the maximum performance before layer $i$ be denoted as:
\begin{equation}
M_i = \max_{j<i} x_j.    
\end{equation}
Based on the condition $x_i > M_i$, we can identify multiple candidate sets $S$ of pruning layers.
\begin{equation}
    S = \left \{ i_1, i_2, ..., i_k \right \} , \quad 3 \le i_1,...\le i_k \le L.
\end{equation}
Ideally, model performance can be expressed as a function of the selected pruning layers.
\begin{equation}
    y(t)=
\begin{cases}
x_2, & 1 \le t < i_1, \\
x_{i_2}, & i_1 \le t < i_2, \\
\;\vdots & \\
x_{i_K}, & i_K \le t \le L .
\end{cases}
\end{equation}
As the impact of visual token selection propagates through subsequent layers, we reformulate layer selection as an optimization problem that maximizes the overall layer contribution under a fixed budget of $m$ pruning layers.

Let $i_{K+1} = L, i_0 = 2$. Then the model performance is formulated as:
\begin{equation}
    P(s) = \frac{\sum_{k=0}^{K} x_{i_k}(i_{k+1} - i_k)}{L-2}.
\end{equation}
Let $U(s)$ denote the integral in the numerator. If the previous update occurs at layer $i_{k-1}$ and the next at layer $j$, then the marginal area contribution of current update $i$ is:
\begin{equation}
    \Delta U(i|i_{k-1}, j) = (x_i - x_{i_{k-1}})(j-i).
\end{equation}
This constitutes a dynamic programming problem. Consider the last update: it can occur either at the current layer $i$ or at a later layer $j$. The necessary and sufficient condition for $j$ to be preferable to $i$ is:
\begin{equation}
    x_j(L-j) \ge x_i(L-i) - x_{i_{m-1}}(j-i).
\end{equation}
This establishes the state transition equation. The optimal solution, and therefore the optimal pruning layers, follows directly.



As shown in Fig.\ref{fig:intro-select}, we conduct layer selection experiments using LLaVA-1.5-7B on three localization datasets (RefCOCO, RefCOCO+, RefCOCOg) and three non-localization datasets (TextVQA, GQA, V2-VQA). From the training split of each dataset, 1,000 instances are randomly sampled for evaluation. 



Despite dataset-specific variations, consistent patterns can still be observed across datasets.
In particular, early layers exhibit noticeable fluctuations, and performance consistently peaks around layer 15, suggesting shared characteristics in layer-wise token discriminability.

Performance metrics are first normalized across all datasets and then averaged to obtain $\left \{ x_i \right \}_{i=1}^L$.
Following the above layer selection protocol, layers 3, 11, and 15 are selected as pruning layers.

\begin{figure*}[t]
  \begin{center}
    \centerline{\includegraphics[width=\textwidth]{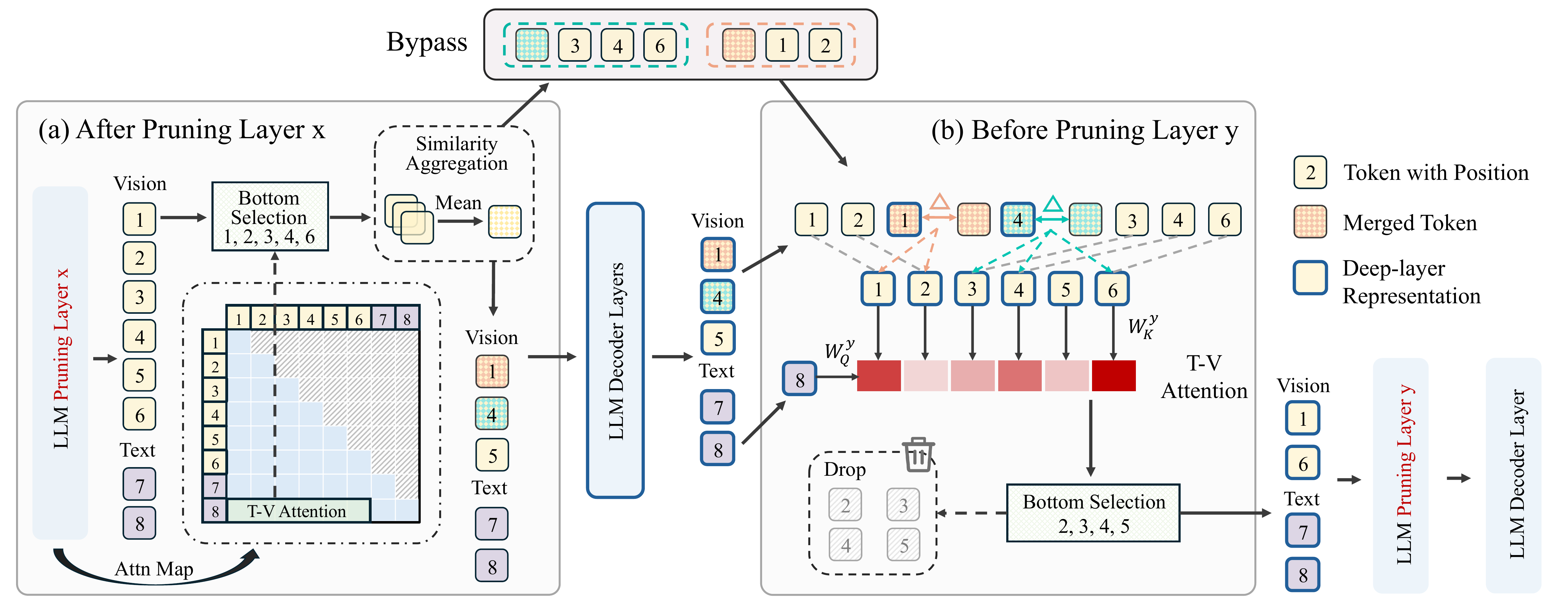}}
    \caption{
    \textbf{SwiftVLM architecture overview.}
    (a) After layer $x$, unselected visual tokens are grouped for bypassing, with the resulting merged tokens participating in subsequent computation.
    (b) Before layer $y$, token alignment is applied to restore grouped tokens, enabling re-evaluation of visual tokens at layers with stronger token selection capability.
    }
    \label{fig:archi}
  \end{center}
\end{figure*}

\subsection{Architecture}

For each model, we first select a set of pruning layers, denoted as layers $x$ and $y$ in Fig.\ref{fig:archi}.

The first pruning operation is performed after layer $x$. Based on the attention map produced by layer $x$, we extract the T–V attention scores between the last text token and all visual tokens. 
The top-ranked visual tokens are retained and directly propagated to layer $x+1$ for further inference.
 The remaining low-ranked visual tokens are grouped according to the similarity between their hidden states, measured by 
\begin{equation}
 s_{i,j} = \frac{\left(\mathbf{h}_i^{\,x}\right)^{\top} \mathbf{h}_j^{\,x}
 }{|\mathbf{h}_i^{\,x}||\mathbf{h}_j^{\,x}|},
 \end{equation}
where $\mathbf{h}_i$ and $\mathbf{h}_j$ denote the hidden states of visual tokens $i$ and $j$, respectively.
Visual tokens within the same group are then merged by averaging their hidden states across feature dimensions, yielding a single merged token
\begin{equation}
\tilde{\mathbf{h}}_{gm}^{\,x} =\tilde{\mathbf{h}}_g^{\,x} =  \frac{1}{|\mathcal{G}_g|} \sum_{i \in \mathcal{G}_g} \mathbf{h}_i^{\,x},
 \end{equation}
which participates in the computation of layer $x+1$.

Here, we propose a new pruning strategy termed \textbf{bypass}. Instead of permanently discarding unselected visual tokens, bypass preserves these tokens and forwards them through a side pathway to the next pruning layer, where they re-participate in the pruning selection process.

Before the pruning layer $y$, we re-evaluate the importance of all visual tokens. For each group formed by merged tokens, we estimate the average offset of the group as
\begin{equation}
\Delta \mathbf{h}_{gm}
= \tilde{\mathbf{h}}_{gm}^{\,y-1}
- \tilde{\mathbf{h}}_{gm}^{\,x}.
\end{equation}

To align the visual tokens transmitted through the bypass pathway with the deeper representations of other tokens, we correct each visual token in group $g$ as follows:
\begin{equation}
\hat{\mathbf{h}}_{i}^{\,y-1}
= \mathbf{h}_i^{\,x} + \Delta \mathbf{h}_{gm},
\qquad i \in \mathcal{G}_g.
\end{equation}

Using the aligned visual tokens and the key projection matrix $W^y_K$ of pruning layer $y$, we construct the key representations. The query is obtained by projecting the last text token from layer $y-1$ with $W^y_Q$. We then compute the T–V attention and perform visual token selection once again.
At this stage, only the selected important visual tokens are retained to participate in the subsequent prefill computation.

\subsection{Representation Alignment Analysis}
\label{sec:representation}
Transformer~\cite{vaswani2017attention} layers adopt a residual formulation, where the hidden states are updated as
\begin{equation}
\mathbf{h}^{\ell}
= \mathbf{h}^{\ell-1} + \mathcal{F}^{\ell}(\mathbf{h}^{\ell-1}),
\end{equation}
with $\mathcal{F}^{\ell}(\cdot)$ denoting the combined attention and feed-forward transformation at layer $\ell$.

For a visual token $i$ belonging to group $\mathcal{G}_g$, its hidden state in the vanilla model evolves from layer $x+1$ to layer $y\!-\!1$ as
\begin{equation}
\mathbf{h}_i^{y-1}
= \mathbf{h}_i^{x}
+ \sum_{\ell=x+1}^{y-1} \mathcal{F}^{\ell}(\mathbf{h}_i^{\ell-1}).
\end{equation}

Taking the average over all tokens in group $\mathcal{G}_g$, we obtain
\begin{equation}
\tilde{\mathbf{h}}_g^{\,y-1}
= \tilde{\mathbf{h}}_g^{\,x}
+ \sum_{\ell=x+1}^{y-1}
\frac{1}{|\mathcal{G}_g|}
\sum_{i \in \mathcal{G}_g}
\mathcal{F}^{\ell}(\mathbf{h}_i^{\ell-1}),
\end{equation}
We denote by $\Delta \mathbf{h}_{g}$ the accumulated group-level residual update. 

In Sec.\ref{sec:exp-why}, we obtain $\Delta \mathbf{h}_{g}$ from the vanilla model and compare it with $\Delta \mathbf{h}_{gm}$. Under fine-grained grouping, their low-dimensional projections show near-complete overlap, providing empirical support for the proposed offset-based approximation.

\begin{table*}[t]
\centering
\small
\setlength{\tabcolsep}{3.5pt}
\renewcommand{\arraystretch}{1.25}

\caption{\textbf{Performance comparison under different visual token budgets.} (+) and (g) denote RefCOCO+ and RefCOCOg, respectively.}
\begin{tabular}{lcccccccccccc}
\Xhline{1pt} 
\multirow{2}{*}{\textbf{Method}} & \multicolumn{4}{c}{\textbf{Localization}} & \multicolumn{7}{c}{\textbf{Non-localization}} &\multirow{2}{*}{\textbf{FLOPs (T)}}\\
\cmidrule(lr){2-5}
\cmidrule(lr){6-12}

&\textbf{RefCOCO} & \textbf{(+)} & \textbf{(g)}  & \textbf{Avg.} &\textbf{VQA}$^{\text{Text}}$ &\textbf{GQA} & \textbf{SQA} & \textbf{MME} &\textbf{MMB}& \textbf{POPE} & \textbf{Avg.}&  \\
\Xhline{1pt} 

\rowcolor{lightgray}
\multicolumn{13}{c}{\textit{Upper Bound, 576 Tokens (100\%)}} \\

\multirow{2}{*}{Vanilla}
& 75.9 & 67.0 & 70.7 & \multirow{2}{*}{100\%} & 46.9 & 61.4 & 69.6 & 1509 & 64.6 & 86.8
& \multirow{2}{*}{100\%} & \multirow{2}{*}{4.29}  \\
& 100\% & 100\% & 100\% && 100\% & 100\% & 100\% & 100\% & 100\%  & 100\%
&  &  \\
\hline

\rowcolor{lightgray}
\multicolumn{13}{c}{\textit{Retain 192 Tokens ($\downarrow$ 66.7\%)}} \\

FastV
& 30.6 & 25.8 & 29.7 & \multirow{2}{*}{40.3\%} & 43.6 & 57.2 & 69.4 & 1471 & 63.2
& 82.0 &\multirow{2}{*}{95.9\%}& \multirow{2}{*}{1.71} \\
\scriptsize\textcolor{gray}{\textit{(ECCV'24)}}
& 40.3\% & 38.5\% & 42.0\% && 93.0\% & 93.2\% & 99.7\% & 97.5\% & 97.8\% & 94.5\%
&  &  \\
\hline

VisionZip
& 7.0 & 5.7 & 6.3 &\multirow{2}{*}{8.9\%} &45.2 & 58.9 & 68.8 &1460 &62.9 
&86.6 &\multirow{2}{*}{97.5\%} & \multirow{2}{*}{1.71} \\
\scriptsize\textcolor{gray}{\textit{(CVPR'25)}}
& 9.2\% & 8.5\% & 8.9\% && 96.4\% & 95.9\% & 98.9\% & 96.8\% & 97.4\%& 99.8\%& &\\
\hline

PDrop
& 22.2 & 18.2 & 18.7 &\multirow{2}{*}{27.6\%}& 42.9 & 55.5 & 69.2 & 1365 & 63.2
&81.1 & \multirow{2}{*}{93.8\%}& \multirow{2}{*}{1.72} \\
\scriptsize\textcolor{gray}{\textit{(CVPR'25)}}
& 29.2\% & 27.2\% & 26.4\% && 91.5\% & 90.4\% & 99.4\% & 90.5\% & 97.8\%&93.4\%
&  &  \\
\hline

SparseVLM
& 8.7 & 7.5 & 7.1 &\multirow{2}{*}{10.9\%} & \textbf{45.8} & 58.9 & 69.1 & 1447 &64.2& 86.7
 &\multirow{2}{*}{98.0\%} & \multirow{2}{*}{1.72} \\
\scriptsize\textcolor{gray}{\textit{(ICML'25)}}
& 11.5\% & 11.2\% & 10.0\% & & \textbf{97.7\%} & 95.9\% & 99.3\% & 95.9\% &99.4\%& 99.9
&  &  \\
\hline

FEATHER
& 52.0 & 45.5 & 45.4 &\multirow{2}{*}{66.9\%} & 42.9& 58.6 & \textbf{70.5} &1431 &63.9 &84.4
&\multirow{2}{*}{96.5\%} & \multirow{2}{*}{1.82} \\
\scriptsize\textcolor{gray}{\textit{(ICCV'25)}}
& 68.5\% & 67.9\% & 64.2\% & &91.5\% & 95.4\% & \textbf{101.3\%} & 94.8\% &98.9\% & 97.2\%
&  &  \\
\hline

\multirow{2}{*}{SwiftVLM}
& \textbf{66.6} & \textbf{58.5} & \textbf{60.6} &\multirow{2}{*}{\textbf{86.9\%}} & 45.3  & \textbf{60.7} & 69.0 & \textbf{1503} & \textbf{64.5}&\textbf{87.1}&\multirow{2}{*}{\textbf{99.0\%}} & \multirow{2}{*}{1.75} \\
& \textbf{87.7\%} & \textbf{87.3\%} & \textbf{85.7\%} & & 96.6\% & \textbf{98.9\%} & 99.1\% & \textbf{99.6\%} & \textbf{99.8\%} & \textbf{100.3\%}
&  &  \\
\hline

\rowcolor{lightgray}
\multicolumn{13}{c}{\textit{Retain 128 Tokens ($\downarrow$ 77.8\%)}} \\

FastV
& 12.8 & 11.1 & 13.8 &\multirow{2}{*}{17.7\%} & 39.7& 53.6 & 68.5 &1377&62.3&77.7&\multirow{2}{*}{91.3\%}& \multirow{2}{*}{1.29} \\
\scriptsize\textcolor{gray}{\textit{(ECCV'24)}}
& 16.9\% & 16.6\% & 19.5\% && 84.6\% & 87.3\% & 98.4\% & 91.3\% &96.4\%& 89.5\%
&  &  \\
\hline

VisionZip
& 4.6 & 3.6 & 4.3 &\multirow{2}{*}{5.8\%} &\textbf{44.4} & 57.5 & 68.9& 1441 &62.0 
& 85.1&\multirow{2}{*}{96.1\%} & \multirow{2}{*}{1.29} \\
\scriptsize\textcolor{gray}{\textit{(CVPR'25)}}
& 6.0\% & 5.4\% & 6.1\% && \textbf{94.7\%} & 93.6\% & 99.0\% & 95.5\% & 96.0\%
&98.0\%  &  &\\
\hline

PDrop
& 3.0 & 2.3 & 2.3&\multirow{2}{*}{3.6\%}  & 39.9 & 54.3 & \textbf{70.2} & 1322 &61.9 & 80.9 &\multirow{2}{*}{91.8\%} & \multirow{2}{*}{1.28} \\
\scriptsize\textcolor{gray}{\textit{(CVPR'25)}}
& 4.0\% & 3.4\% & 3.3\% && 85.1\% & 88.4\% & \textbf{100.9\%} & 87.6\% &95.8\%& 93.2\%
&  &  \\
\hline

SparseVLM
& 4.8 & 3.9 & 4.1 &\multirow{2}{*}{6.0\%} & 42.0 & 57.4 & 69.8 & 1418 &
63.6& 86.0 & \multirow{2}{*}{95.8\%} &\multirow{2}{*}{1.30}\\
\scriptsize\textcolor{gray}{\textit{(ICML'25)}}
& 6.3\% & 5.8\% & 5.8\%& & 89.6\% & 93.5\% & 100.3\% & 94.0\% &98.5\%& 99.1\%
&  &  \\
\hline

FEATHER
&39.0& 34.3 & 35.2 & \multirow{2}{*}{50.8\%} & 41.2 & 56.5 &69.6  &1453  & 63.2
&83.3 &\multirow{2}{*}{95.0\%} & \multirow{2}{*}{1.44}\\
\scriptsize\textcolor{gray}{\textit{(ICCV'25)}}
&51.4\% & 51.2\% & 49.8\%  & & 87.8\% & 92.0\% & 100\% &96.3\% & 97.8\%
& 96.0\% &  \\
\hline


\multirow{2}{*}{SwiftVLM}
& \textbf{55.2}&\textbf{46.6}  & \textbf{47.4} & \multirow{2}{*}{\textbf{69.8\%}} & 41.8 & \textbf{59.2} & 68.5 & \textbf{1477} & \textbf{63.9} & \textbf{86.1} &\multirow{2}{*}{\textbf{96.7\%}}& \multirow{2}{*}{1.31} \\
& \textbf{72.7}\% & \textbf{69.6\%} & \textbf{67.0\%} & & 89.1\% & \textbf{96.4\%} & 98.4\% & \textbf{97.9\%} & \textbf{98.9\%}& \textbf{99.2\%} &&  \\

\bottomrule
\end{tabular}

\label{tab:exp-overall}
\end{table*}

\subsection{FLOPs Computation}

We consider a setting where visual tokens are pruned after the $K$-th VLM layer, removing a fraction $D\%$ of visual tokens. Let $n_v$ and $n_t$ denote the numbers of visual tokens and non-visual tokens, respectively, with $T$ layers, hidden dimension $d$, and FFN intermediate dimension $m$. The total number of tokens is $n = n_v + n_t$. and the token count after pruning becomes $\hat{n} =  (1 - D\%) * n_v + n_t$. The resulting FLOPs $F$ are given by:
\begin{align}
    C_n &= \bigl(4nd^2 + 2n^2d + 3ndm\bigr),\\
    F &= K \times C_n + (T-K) \times C_{\hat{n}}.
\end{align}

Furthermore, we analyze the additional computational overhead introduced by the proposed operation. Let $R$ denote the number of low-ranked visual tokens and $Z$ the number of merged tokens. 

The merge step incurs an overhead of $2RZd$.
Representation alignment adds an extra cost of $R d$.
Projecting the last text token to form the query costs $2d^2$, while projecting the visual tokens and computing the subsequent dot products introduce costs of $2n_v d^2$ and $2n_v d$, respectively.
Let $r$ denote the ratio of visual tokens retained at layer $y$.
The overall computational overhead $F_o$ is thus given by
\begin{equation}
\mathrm{F_o}
=
2RZd+Rd
+ 2n_v d+2d^2
+ 2(1-r)n_v d^2 .
\end{equation}
\section{Experiments}
\subsection{Overall Performance}
\textbf{Datasets.} We categorize inference tasks into localization and non-localization types, where the former emphasizes fine-grained visual details and the latter focuses on holistic information integration.
We evaluate our method on nine widely used benchmarks, including RefCOCO, RefCOCO+, RefCOCOg~\cite{kazemzadeh2014referitgame, yu2016modeling}, TextVQA, GQA~\cite{hudson2019gqa}, SQA~\cite{lu2022learn}, MME~\cite{bolya2022token}, MMB~\cite{liu2024mmbench}, POPE~\cite{li2024seed}. For TextVQA, we follow prior work~\cite{endo2025feather} and exclude OCR prompt to better evaluate how pruning affects visual understanding.

\textbf{Main Results.} Since the average RefCOCO bounding box covers about 102 visual tokens, Tab.~\ref{tab:exp-overall} reports the performance of different methods on LLaVA-1.5-7B under two visual token budgets (192 and 128). Across non-localization tasks, all methods achieve competitive performance, including VisionZip, which employs text-agnostic feature compression.

In contrast, performance differences become pronounced on localization tasks. Notably, PDrop and SparseVLM do not preserve the positional information of visual tokens after pruning, leading to substantial performance degradation~\cite{chien2025grounding}.
FEATHER mitigates the impact of RoPE by recomputing attention, resulting in higher FLOPs compared to other methods.
Moreover, despite eliminating RoPE effects, the ability of different layers to discriminate important visual tokens in FEATHER remains non-monotonic, and low-ranked visual tokens are still dropped after the initial pruning stage. As a result, FEATHER underperforms SwiftVLM by by roughly 20\%.

\begin{figure}
  \begin{center}
    \centerline{\includegraphics[width=\columnwidth]{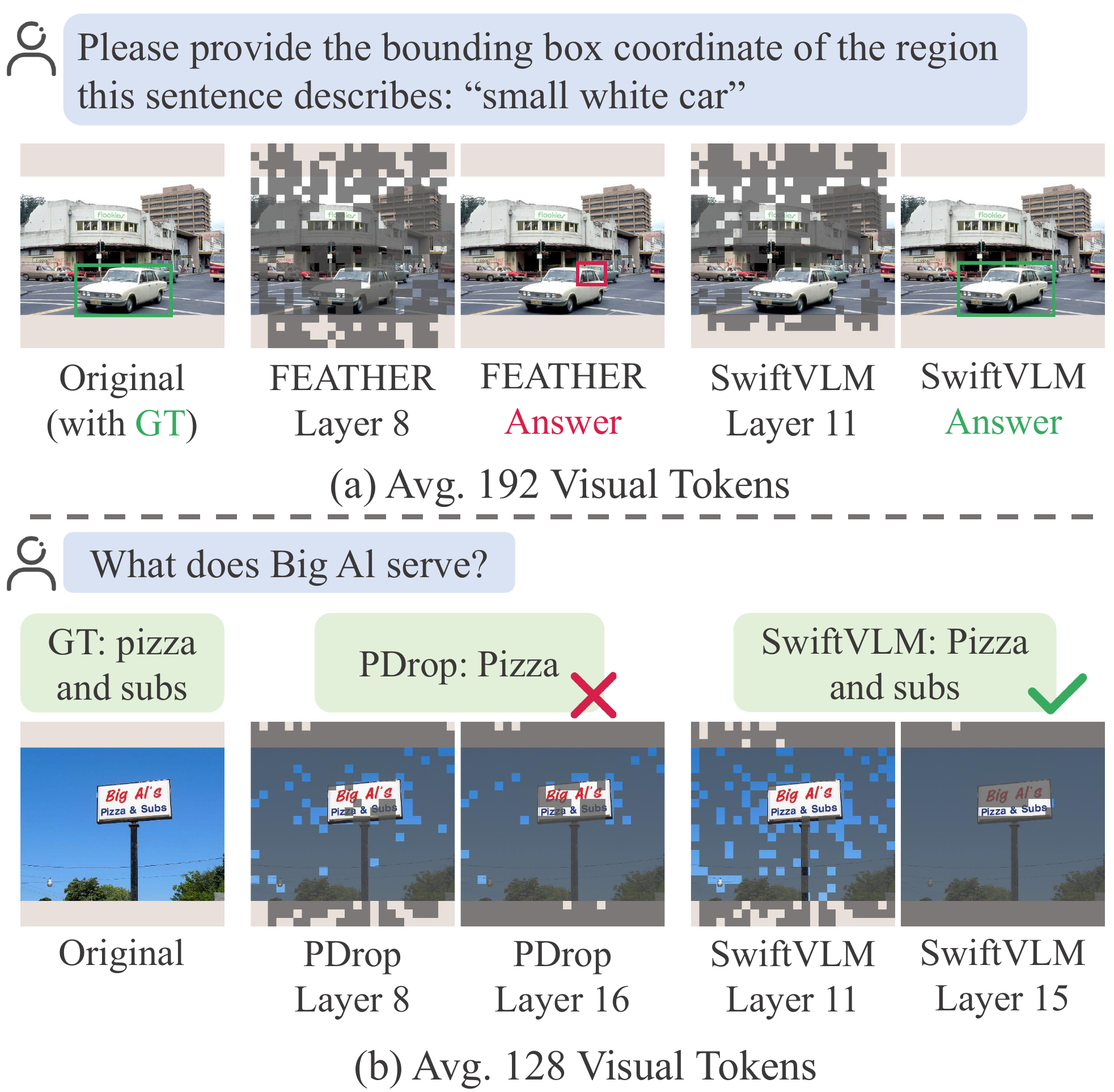}}
     \caption{\textbf{Visualization of method performance under varying tasks and computation budgets.}}
    \label{fig:exp-visual}
  \end{center}
\end{figure}

\textbf{Visualization.} We visualize examples from RefCOCO and TextVQA, showing the retained visual tokens as image patches along with the final answers. As illustrated in Fig.\ref{fig:exp-visual}, FEATHER and PDrop adopt drop-based pruning and discard task-relevant visual tokens (e.g., the car in localization and the signboard in VQA), leading to incomplete or incorrect answers.

\subsection{Efficiency Study}

\begin{table}[t]
\centering
\small
\setlength{\tabcolsep}{3pt}
\renewcommand{\arraystretch}{1.4}

\caption{\textbf{Efficiency study on LLaVA-1.5-7B.} Total Time denotes the wall-clock time required to process the entire POPE dataset. Prefilling Time refers to the average prefill latency per sample. $\Delta$ indicates the speedup factor relative to the vanilla model.}
\begin{tabular}{c|ccccc}
\Xhline{1pt} 
\multirow{2}{*}{\textbf{Tokens}} & \multirow{2}{*}{\textbf{Method}} & \multirow{2}{*}{\makecell{\textbf{Total}\\\textbf{Time (s)}}}
  & \multirow{2}{*}{\textbf{$\Delta$ }} & \multirow{2}{*}{\makecell{\textbf{Prefilling}\\\textbf{Time (ms)}}}
&\multirow{2}{*}{\textbf{$\Delta$}} \\
& & & & &\\
\Xhline{1pt} 
576 & Vanilla & 850.7 & - & 67.3 & -\\

\hline

\multirow{3}{*}{192}  &FastV & 551.8& 1.54$\times $ & 34.7 & 1.92$\times $\\
&SparseVLM & 612.3 & 1.39$\times $ & 40.7 & 1.65$\times $\\
&SwiftVLM & 573.8 & 1.48$\times $ & 37.6 & 1.79$\times $ \\

\hline

\multirow{3}{*}{128}  &FastV & 539.4& 1.58$\times $ & 32.8 & 2.05$\times $\\
&SparseVLM & 583.9 & 1.46$\times $ & 37.5 & 1.79$\times $\\
&SwiftVLM & 546.2 & 1.56$\times $ & 33.0 & 2.04$\times $\\

\bottomrule
\end{tabular}

\label{tab:exp-delay}
\end{table}

Following SparseVLM, we implement SwiftVLM in a FlashAttention-compatible~\cite{dao2022flashattention} manner and report the corresponding latency results in Tab.\ref{tab:exp-delay}. Compared to the vanilla model, all pruning-based methods achieve noticeable speedups.
FastV attains the largest acceleration since it performs pruning only once.

Unlike FLOPs computation, FlashAttention does not provide direct access to attention maps, requiring attention scores to be recomputed in practice.
Consequently, SwiftVLM incurs lower latency than SparseVLM, as it only computes attention between the final text token and visual tokens, whereas SparseVLM requires attention computation for all text tokens.

\begin{table}[t]
\centering
\small
\setlength{\tabcolsep}{5pt}
\renewcommand{\arraystretch}{1.2}

\caption{\textbf{Ablation study.} $\mathrm{X_S}$ denotes layer selection. $\mathrm{X_M}$ denotes token merging, and $\mathrm{X_B}$ denotes the bypass mechanism.}
\begin{tabular}{c|lcc}
\Xhline{1pt} 
\textbf{Tokens} & \textbf{Method} & \textbf{RefCOCO} & \textbf{VQA}$^{\text{Text}}$ \\
\Xhline{1pt} 

\multirow{4}{*}{192}  &Baseline & 42.6 & 43.2\\

& + \textsc{X}$_\textsc{S}$ & 64.5 & 45.3 \\

& + \textsc{X}$_\textsc{S}$ + \textsc{X}$_\textsc{M}$ & 63.7 & 44.8 \\

& \cellcolor{lightgray}\textbf{+ \textsc{X}$_\textsc{S}$ + \textsc{X}$_\textsc{M}$ + \textsc{X}$_\textsc{B}$} & \cellcolor{lightgray}\textbf{66.6} & \cellcolor{lightgray}\textbf{45.3} \\

\hline

\multirow{4}{*}{128} &Baseline & 23.2 & 41.2  \\

& + \textsc{X}$_\textsc{S}$ & 42.8 & 40.1 \\

& + \textsc{X}$_\textsc{S}$ + \textsc{X}$_\textsc{M} $ & 51.9 & 40.7 \\

&\cellcolor{lightgray}\textbf{+ \textsc{X}$_\textsc{S}$ + \textsc{X}$_\textsc{M}$ + \textsc{X}$_\textsc{B}$} & \cellcolor{lightgray}\textbf{55.2} & \cellcolor{lightgray}\textbf{41.8} \\

\bottomrule
\end{tabular}

\label{tab:exp-ab}
\end{table}














\begin{figure}[t]
  \begin{center}
    \centerline{\includegraphics[width=\columnwidth]{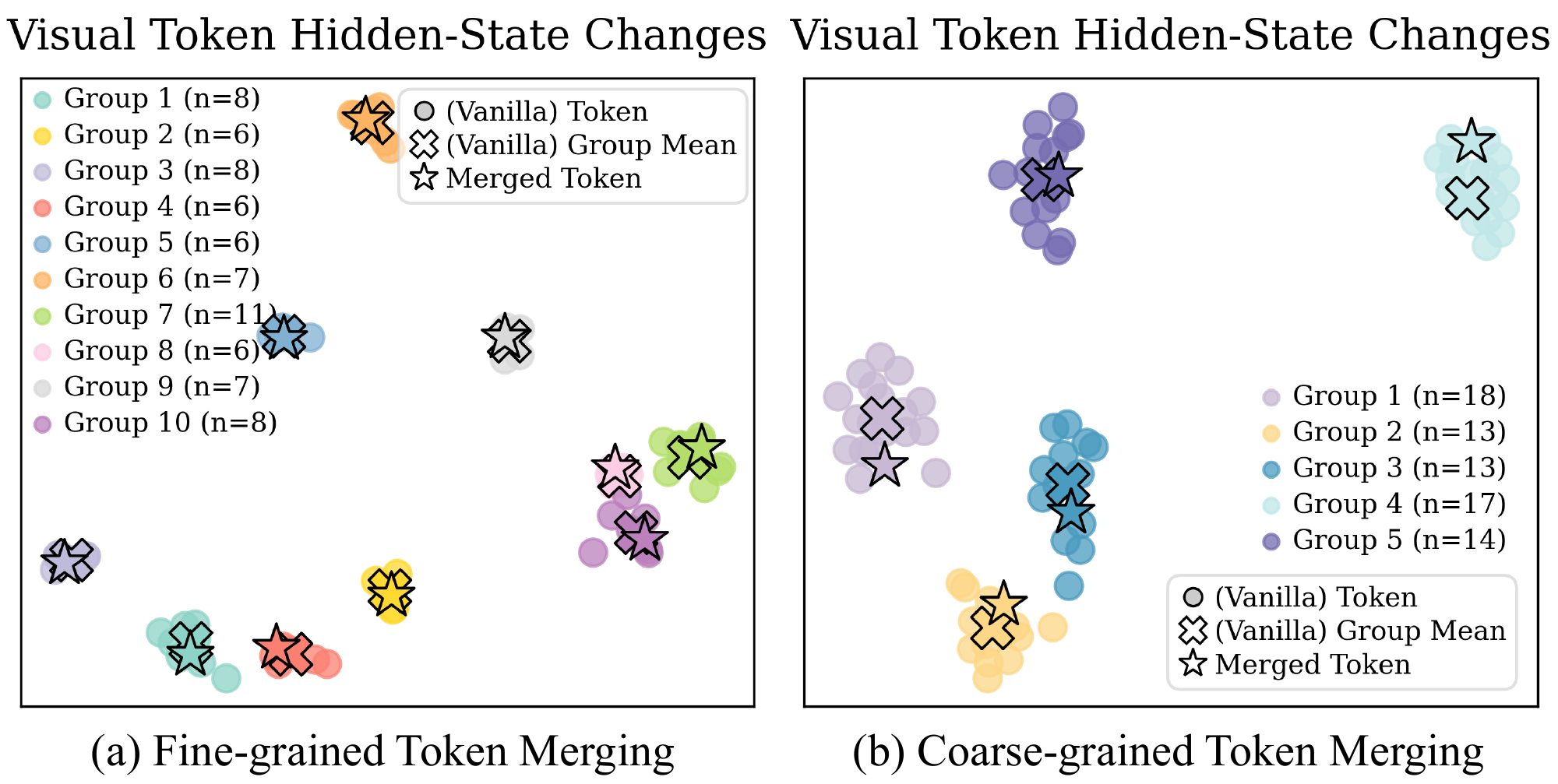}}
    \caption{
      \textbf{t-SNE visualization of visual token hidden-state changes.} Colors denote similarity-based token groups. In the vanilla model, $\bullet$ shows per-token changes and $\times$ shows the group-wise mean. In our method, each group is merged into a single token, its change from layer 3 to layer 10 is shown as a $\bigstar$. At $n=18$, merged tokens account for less than 5\%.}
    \label{fig:exp-tsne}
  \end{center}
\end{figure}

\subsection{Ablation Study}

We adopt PDrop as the baseline and augment it with positional encoding updates. Based on this configuration, we progressively introduce layer selection, token merging, and bypass, with results reported in Tab.~\ref{tab:exp-ab}.

Under the 192-token setting, pruning at layers with monotonically increasing selection capability yields the largest gains, while token merging degrades performance due to unnecessary information compression under sufficient computation budget. In contrast, under the more constrained 128-token setting, token merging becomes beneficial, as aggressive dropping would otherwise remove critical visual information. Overall, pruning with bypass consistently provides stable performance improvements across different budget settings.

\begin{figure}[t]
  \begin{center}
    \centerline{\includegraphics[width=\columnwidth]{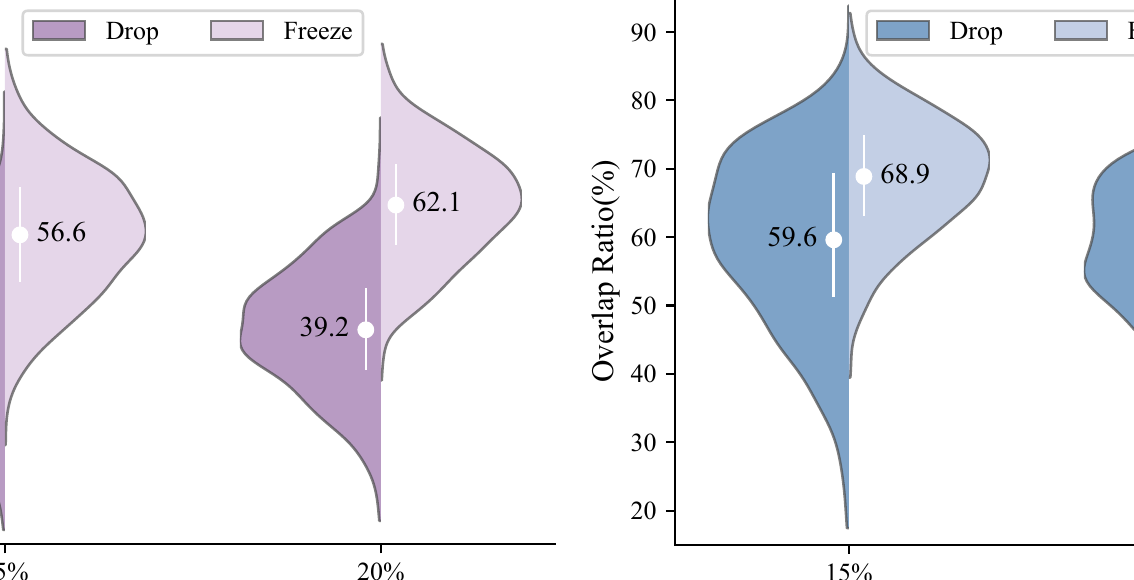}}
    \caption{
      \textbf{Token selection overlap with vanilla for drop and bypass.} Under an equal computational budget, the overlap distribution and mean are reported over 4,000 cases by comparing the tokens selected at layer 15 under different pruning schemes with those selected by the vanilla model, in order to assess their impact on intrinsic selection behavior.}
    \label{fig:exp-overlap}
  \end{center}
\end{figure}

\subsection{Why Bypass Works?}
\label{sec:exp-why}

To investigate why visual tokens forwarded through bypass can still participate effectively in subsequent computation after representation alignment, we analyze the low-dimensional projections of token offsets as described in Sec.\ref{sec:representation}. Under the 128-token setting, we visualize the results for a sample in TextVQA, as shown in the Fig.\ref{fig:exp-tsne}(a). Here, Merged Token corresponds to the offset $\Delta h_{gm}$. For each bypassed group, we additionally run the vanilla model.
Vanilla Token records the actual hidden-state changes of individual tokens within the group after layer 10, while Vanilla Group Mean represents the average hidden-state change computed from these tokens. We observe that the vanilla group mean closely overlaps with the merged token offset and remains highly consistent with the changes of individual tokens within the group. We then substantially reduce the number of merged tokens and report the results for the same example in Fig.\ref{fig:exp-tsne}(b).

Given that VLMs employ causal attention, the hidden-state evolution of a visual token can actually only be influenced by preceding visual tokens. Moreover, since attention fundamentally operates through similarity-based interactions, we hypothesize that visual tokens with similar semantics exhibit similar transformation directions in the representation space, and can thus be well approximated by the changes of the corresponding merged token.

\subsection{Why Is Bypass Better Than Drop?}

Under the 128-token setting, we compare the visual tokens retained at layer 15 by drop and bypass with the top 5\% and top 10\% tokens selected by the vanilla model, and report their overlap ratios on TextVQA and RefCOCO in Fig.\ref{fig:exp-overlap}.

Bypass exhibits a higher overlap with the vanilla model, indicating its ability to preserve visual tokens that are critical for reasoning.
This overlap gap is more pronounced on RefCOCO, consistent with the larger performance differences observed across datasets under the 128-token setting in the ablation study.

\subsection{Generalization}

\begin{table}[t]
\centering
\small
\setlength{\tabcolsep}{3pt}
\renewcommand{\arraystretch}{1.3}

\caption{\textbf{Performance comparison on LLaVA-NeXT-7B.}}

\definecolor{rowgray}{gray}{0.90}
\definecolor{mygreen}{RGB}{67,160,71}

\begin{tabular}{lccccc}
\noalign{\hrule height 1.2pt}
\bfseries Method & \bfseries RefCOCO & \bfseries VQA$^{\text{Text}}$ & \bfseries GQA & \bfseries MMB & \bfseries Rel.\ Acc \\
\hline

\rowcolor{rowgray}
\multicolumn{6}{c}{\rule{0pt}{2.8ex}\itshape Upper Bound, Retain 100\% Tokens } \\

Vanilla   & 85.3 & 65.5 & 63.9 & 67.9 & 100\% \\

\rowcolor{rowgray}
\multicolumn{6}{c}{\rule{0pt}{2.8ex}\itshape Retain 33.3\% Tokens} \\

FastV     & 40.5 & 58.7 & 59.0 & 48.3 & 75.1\% \\
FEATHER   & 68.8 & 62.6 & 62.5 & 67.5 & 92.8\% \\
SwiftVLM & \bfseries 80.7 & \bfseries 64.1 & \bfseries 63.6 & \bfseries 68.0 & \bfseries 98.0\% \\

\rowcolor{rowgray}
\multicolumn{6}{c}{\rule{0pt}{2.8ex}\itshape Retain 22.2\% Tokens} \\

FastV     & 26.1 & 52.6 & 56.9 & 46.0 & 66.9\% \\
FEATHER   & 53.1 & 60.9 & 61.9 & 66.5 & 87.5\% \\
SwiftVLM & \bfseries 79.6 & \bfseries 62.4 & \bfseries 63.5 & \bfseries 67.7 & \bfseries 97.1\% \\

\noalign{\hrule height 1.2pt}
\end{tabular}
\end{table}

To evaluate generalization, following prior work, we conduct experiments on LLaVA-NeXT~\cite{liu2024llavanext} across four datasets.
Due to image padding removal in LLaVA-NeXT, performance is compared using visual token retention ratios.
SwiftVLM consistently outperforms other methods, with particularly notable gains on localization datasets.
\section{Conclusion}
In this work, we revisit visual token pruning in VLMs and reveal that visual token importance varies substantially across layers. This observation explains why existing drop-based pruning methods, which rely on early selection decisions, often struggle on tasks requiring fine-grained visual reasoning.
To better preserve visual information, we introduce a novel pruning strategy, termed bypass, and integrate it into our proposed pruning framework, SwiftVLM. This design allows each pruning layer to perform token selection in a relatively independent manner.
Experimental results demonstrate that bypass consistently outperforms drop, suggesting its potential as a promising pruning paradigm.

\bibliography{example_paper}
\bibliographystyle{icml2026}

\end{document}